\documentclass{article}

\PassOptionsToPackage{numbers,square,sort&compress}{natbib}
\usepackage[nonanonymous]{neurips_2026}
\makeatletter
\renewcommand{\@notice}{}
\makeatother

\usepackage{float} 
\usepackage{graphicx}
\usepackage{amsmath}
\usepackage{amssymb}
\usepackage{makecell}
\usepackage{algorithm}
\usepackage{algorithmic}
\newtheorem{proposition}{Proposition}
\newtheorem{lemma}{Lemma}
\usepackage{subcaption}
\usepackage{multirow}

\newtheorem{corollary}{Corollary}

\newcommand{\algthickrule}{\hrule height 0.8pt}
\newcommand{\algthinrule}{\hrule height 0.4pt}

\usepackage[utf8]{inputenc} 
\usepackage[T1]{fontenc}    
\usepackage{hyperref}       
\usepackage{url}            
\usepackage{booktabs}       
\usepackage{amsfonts}       
\usepackage{nicefrac}       
\usepackage{microtype}      
\usepackage{xcolor}         

\title{Probabilistic Robustness-driven Universal Adversarial Perturbations with Explainability \\against Deep Reinforcement Learning-based \\ Intrusion Detection System}

\author{%
  Hongsen~Zhang\\
  WMG\\
  University of Warwick\\
  Coventry, CV4 7AL \\
  \texttt{Hongsen.Zhang@warwick.ac.uk} \\
  \And
  Lu~Zhang \\
  Wolfson School of Mechanical, Electrical and Manufacturing Engineering \\
  Loughborough University\\
  Loughborough, LE11 3TU\\
  \texttt{L.Zhang12@lboro.ac.uk} \\
  \AND
  Mingjing~Xu \\
  Faculty of Science and Engineering\\
  Swansea university\\
  Swansea, SA2 8PP\\
  \texttt{2596785@swansea.ac.uk} \\
  \And
  Yi~Zhang \\
  WMG\\
  University of Warwick\\
  Coventry, CV4 7AL \\
  \texttt{Yi.Zhang.16@warwick.ac.uk} \\
  \And
  Gregory~Epiphaniou \\
  WMG\\
  University of Warwick\\
  Coventry, CV4 7AL \\
  \texttt{Gregory.Epiphaniou@warwick.ac.uk} \\
  \And
  Carsten~Maple \\
  WMG\\
  University of Warwick\\
  Coventry, CV4 7AL \\
  \texttt{CM@warwick.ac.uk} \\
}

\begin{document}
\nolinenumbers

\maketitle

\begin{abstract}
Deep reinforcement learning (DRL) enables adaptive intrusion detection in dynamic network environments but also exposes intrusion detection systems (IDS) to adversarial threats such as universal adversarial perturbations (UAPs), which apply a single input-agnostic perturbation to degrade detection performance across traffic. Probabilistic Robustness (PR), as a post-hoc evaluation metric, provides a principled, population-level measure of adversarial impact that conceptually aligns with the universality objective of UAPs, i.e., PR quantifies the prevalence of misclassification in the input space, making it a natural signal for guiding UAP generation. Hence, we propose PR-based UAP, which represents the first integration of an explicit PR-driven objective into generating UAPs against DRL-based IDS. Building on this formulation, we introduce PX-UAP, which leverages explainable artificial intelligence (XAI) to guide perturbation shaping under realistic domain constraints, and provides a rigorous theoretical analysis of its design. Extensive experiments demonstrate that PX-UAP consistently outperforms state-of-the-art UAP methods in attack effectiveness.
\end{abstract}

\section{Introduction}

Deep learning (DL) has been extensively applied to intrusion detection systems (IDS) to model high-dimensional network-flow features and improve detection performance~\citep{khraisat2019survey, alkharman2024cyber}. More recently, deep reinforcement learning (DRL) has emerged as an alternative paradigm, enabling IDS to learn adaptive decision policies through interaction-driven reward feedback and to better cope with dynamic network environments and evolving attack patterns~\cite{sutton1998reinforcement,lopez2020application}. Despite these advantages, DL-based IDS are susceptible to adversarial attacks, where carefully crafted input perturbations can induce erroneous predictions~\cite{goodfellow2014explaining}. For DRL-based IDS, this vulnerability has been empirically demonstrated in prior work by applying adversarial examples (AEs), exposing critical security risks at inference time~\cite{zhang2023deep}. In a typical deployment (Fig.~\ref{Figure0}), an IDS continuously monitors network flows and classifies each flow as benign or malicious. An adversary can generate adversarial perturbations by manipulating observed traffic through practical actions, such as packet padding, packet fragmentation, traffic injection, or transmission-timing adjustments, thereby perturbing flow-level features and inducing misclassifications~\cite{debicha2023adv}.

Universal adversarial perturbations (UAPs) constitute a practical and severe threat, by using a single input-agnostic perturbation to induce misclassification across a large fraction of samples, in contrast to instance-specific attacks that require per-sample optimization~\cite{moosavi2017universal}. In intrusion detection, a UAP corresponds to a universal flow-level manipulation applied at inference time, enabling an adversary to consistently degrade detection performance across observed traffic. This universality makes UAP attacks even more serious in real-world IDS deployments, as they can be precomputed and applied efficiently across traffic samples to compromise IDS even under latency constraints and limited attacker resources~\cite{zhang2024universal, zhang2025novel}.





Probabilistic Robustness (PR) has been proposed as a principled framework for characterizing robustness properties of DL models by quantifying the probability that a model’s prediction remains invariant within a specified perturbation region~\cite{webb2018statistical}. Unlike conventional adversarial robustness that focuses on worst-case guarantees, PR provides a distribution-aware, population-level perspective on adversarial impact. Recent studies incorporate PR into adversarial-learning pipelines in the computer vision area, providing instance-specific AE analysis~\cite{karim2023gradient,zhang2024protip} and leveraging PR as a post-hoc selection or evaluation criterion in adversarial training~\cite{zhao2025probabilistic, zhangyi2025adversarial}. 
Importantly, PR can capture adversarial effectiveness as a probability over the data distribution rather than on individual samples. This conceptually aligns with the objective of UAPs, i.e., both quantify the portion of misclassification in the input space: PR focuses on a neighborhood of a given input and UAPs target the whole input space, which makes PR a natural signal for guiding UAP generation. Hence, for the first time, we propose an explicit PR-driven objective for generating effective UAP attacks against DRL-based IDS.

Explainable artificial intelligence (XAI)~\cite{ribeiro2016should} provides feature-level attributions that reveal how DL models make decisions and has been adopted in IDS to analyze feature relevance~\cite{marino2018adversarial,zhang2025explainable}. In adversarial settings, such attributions can guide perturbations toward the most influential features under a limited adversarial budget. However, XAI-guided adversarial attacks remain largely unexplored for UAPs, particularly in IDS scenarios. Hence, in this work, to enhance the attack performance of the proposed PR-based UAP, we further advance XAI to guide the perturbation refinement in UAP generation.
We also provide a theoretical analysis of PX-UAP, showing that the refinement admits an entropy-regularized perturbation budget allocation interpretation and is theoretically aligned with the targeted universal evasion objective.
In summary, the key contributions of this work include:\\
\hspace*{1em}$\bullet$ We propose a Probabilistic Robustness–driven Universal Adversarial Perturbation (\textbf{PR-based UAP}) framework for attacking DRL-based IDS, introducing the concept of PR into UAP generation for the first time.\\
\hspace*{1em}$\bullet$ Building on the PR-based UAP framework, we further enhance attack performance by leveraging XAI, denoted as \textbf{PX-UAP}, which is the first approach to use feature attributions to shape a universal perturbation attack strategy in IDS.\\
\hspace*{1em}$\bullet$ We provide a rigorous theoretical justification for PX-UAP, by establishing a constrained universal evasion objective and an attribution-guided, entropy-regularized perturbation scheme with a first-order improvement guarantee.\\
\hspace*{1em}$\bullet$ We conduct extensive experiments demonstrating that PX-UAP consistently outperforms state-of-the-art UAP methods for intrusion detection in terms of attack effectiveness.

\section{Related Works}
\subsection{Adversarial Attack on Network IDS}
In network IDS applications, network traffic is modeled as flow-based samples, each representing a network connection with temporal behaviors aggregated into statistical features~\cite{moustafa2015unsw}. Adversaries can induce IDS misclassification by perturbing such features at inference time, thereby undermining the reliability of IDS predictions~\cite{zhang2022attention,zhanglu2025vision}. Beyond instance-specific attacks (e.g., FGSM, PGD)~\cite{pacheco2021adversarial,ali2025white}, recent works also demonstrate IDS vulnerability to input-agnostic UAPs~\cite{sheatsley2022adversarial,zhang2024universal}. However, these two UAP-on-IDS researches do not consider realistic domain constraints, including protocol compliance, valid feature ranges, and inter-feature mathematical dependencies~\cite{teuffenbach2020subverting, debicha2023adv}. Zhang et al.~\cite{zhang2025novel} address this gap by integrating inter-feature relationships with feature grouping to maintain validity during UAP generation. Despite the above progress and the inherent universality target of UAPs, no existing work explicitly incorporates a distribution-level objective into the UAP design for a principled generalizable adversarial impact across the distribution.
\subsection{Probabilistic Robustness}
PR is a distribution-aware robustness notion defined over a bounded perturbation region~\cite{webb2018statistical}.
In contrast to worst-case robustness, which considers the \textbf{maximum} classification loss of any AE in the region, PR quantifies the \textbf{probability} that the model’s prediction remains unchanged over the region~\cite{zhao2025probabilistic}. The concept of PR was first introduced by Webb et al.~\cite{webb2018statistical} for AE detection in a white-box threat model and later evolved into a principled framework for robustness evaluation, with extensions to black-box settings~\cite{karim2023gradient,zhang2022proa} and broader multimodal tasks~\cite{zhang2024protip}. However, these works are limited to using PR for robustness assessment. More recently, Zhang et al.~\cite{zhangyi2025adversarial} for\, the\, first time combined PR with adversarial attacks to evaluate the attack effectiveness of candidate AEs for adversarial training. To the best of our knowledge, despite the conceptual compatibility between PR’s distribution-aware characterization of adversarial impact and UAPs’ universality objective, existing research has used PR only in a post-hoc manner to assess robustness or attack performance, leaving PR-driven UAP design unexplored.
\vskip -0.10in
\begin{figure}[H]
  \begin{center}    \centerline{\includegraphics[width=0.84\textwidth]{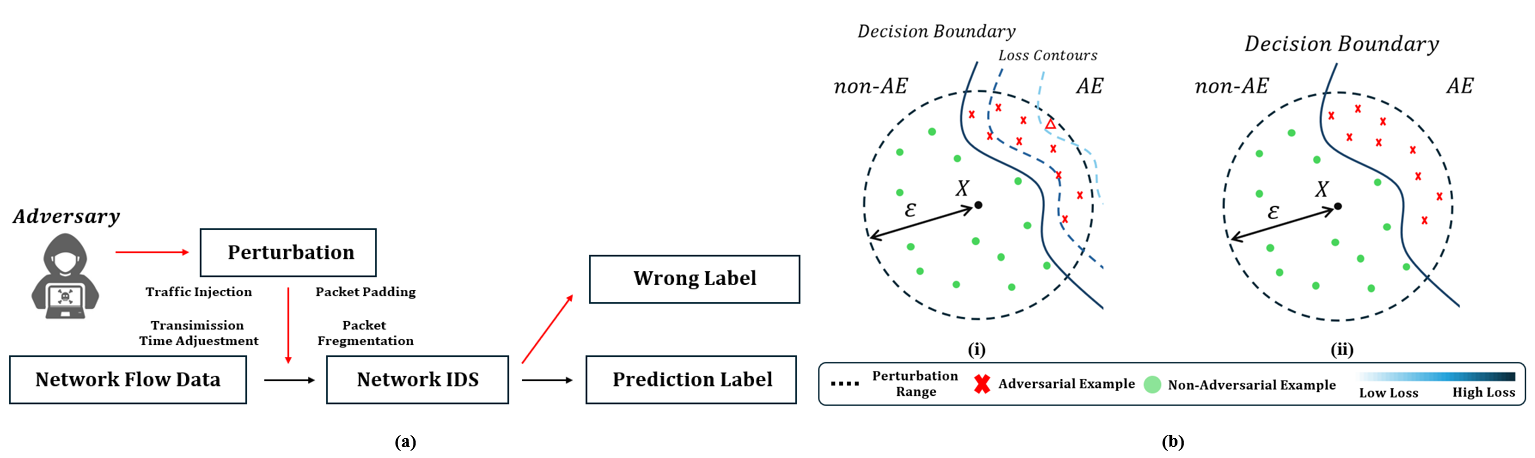}}
    \caption{(a): Adversarial scenario for a network-flow-based IDS. (b): Comparison between worst-case robustness (i) and probabilistic robustness (ii), adapted from~\cite{zhangyi2025adversarial}.
    }
    \label{Figure0}
  \end{center}
  \vskip -0.30in
\end{figure}

\section{Methodology}
In this section, we describe the proposed PR-based UAP, which is the first attempt to integrate PR into UAP generation against DRL-based IDS under realistic domain constraints. We first outline the rationale and intuition behind the proposed PR-based UAP, and then present the detailed attack framework and algorithm. Building upon the PR-based UAP, we introduce PX-UAP, which leverages XAI signals to guide perturbation shaping and further enhance attack performance, representing the first XAI-guided UAP approach in IDS. Furthermore, we provide a rigorous theoretical justification to substantiate the effectiveness of the proposed attack method.
\subsection{Threat model}
In this work, the IDS task is formulated as a binary benign–malicious classification problem. We consider a \textbf{white-box} attacker who has full access to the deployed IDS model (architecture, parameters, and gradients) and the feature preprocessing pipeline for the publicly available CICIDS2018 dataset~\cite{sharafaldin2018toward}, seeking a targeted evasion attack that maps \textbf{malicious} flows to the \textbf{benign} decision. The attacker is constrained to \textbf{$L_2$-bounded} feature-space perturbations with realistic domain constraints. Specifically, following Zhang et al.~\cite{zhang2025novel}, we partition features into Modified Features (MF), Related Features (RF), and Unmodified Features (UF): only MF can be perturbed within feasible feature ranges, then the values of RF are recalculated based on the modified MF, and the UF remain unchanged. This MF-RF recalculation rule is enforced throughout this work. More details are provided in Appx.~\ref{appendix:C}.

\subsection{Rationale behind the proposed PR-based UAP}
\label{sec:PR-rationale}
Given an input $x$ with the ground-truth label $y$, we consider perturbations $\delta \sim \Pr(\cdot \mid x)$ constrained within an $L_p$-norm ball of radius $\epsilon$ (i.e., $\|\delta\|_p \le \epsilon$), yielding the AE $x_{\mathrm{adv}} = x+\delta$. PR is defined as follows~\cite{zhao2025probabilistic}:
\begin{equation}
\label{PR_def}
\mathrm{PR}(x,\epsilon)
=
\mathbb{E}_{\substack{
\delta \sim \Pr(\cdot \mid x) \\
\|\delta\| \le \epsilon
}}
\!\left[
\mathbb{I}_{\{ f_{\theta}(x+\delta)=y \}}(x+\delta)
\right]
\end{equation}
where the indicator $\mathbb{I}_{\{ f_{\theta}(x+\delta)=y \}}$  equals $1$ if classifier $f_{\theta}$ predicts ground truth label $y$ for $x_{\mathrm{adv}}$, and $0$ \\otherwise. Intuitively, in contrast to worst-case robustness (Fig.~\ref{Figure0}.b(i)), $\mathrm{PR}(x,\epsilon)$ measures the fraction of the perturbation region for which $f_{\theta}(x+\delta)=y$, with perturbation distribution $\Pr(\cdot \mid x)$~\cite{karim2023gradient, zhangyi2025adversarial}, as illustrated in Fig.~\ref{Figure0}.b(ii). Thus, a smaller $\mathrm{PR}(x,\epsilon)$ implies that more data samples are misclassified, indicating a larger AE-dominated $\epsilon$-bounded neighborhood of $x$. 
Motivated by this interpretation, prior studies~\cite{zhao2025probabilistic,zhangyi2025adversarial} employ a PR-based variable $k$ as a post-hoc criterion to select the most effective AE among multiple candidates generated by Projected Gradient Descent attack~\cite{madry2017towards}, as formalized in Eq.~\ref{equation:PR_k}:
\begin{equation}
\label{equation:PR_k}
k \;:\; \mathrm{PR}(x+\delta, k)=0,
\quad \forall\, \delta \in \mathrm{PGD}(x),\ \|\delta\|\le\epsilon .
\end{equation}

\begin{figure}[H]
  \begin{center}    \centerline{\includegraphics[width=0.85\textwidth]{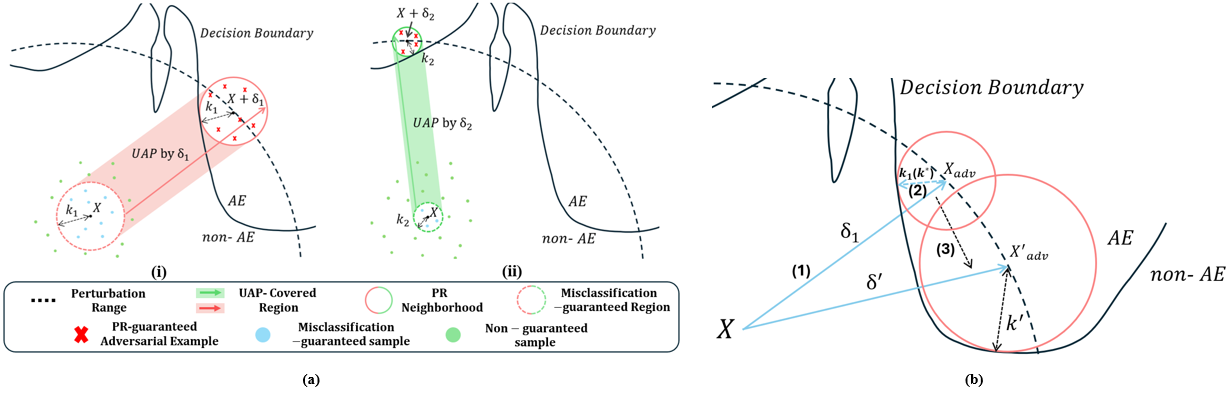}}
    \caption{\textbf{(a)}: Illustration of mapping between PR-based variable $k$ and Misclassification-Guaranteed Region in the UAP setting. Two examples $\delta_1$ and $\delta_2$ are provided in (i) and (ii), respectively. UAP by $\delta_1$ (larger k) induces a larger size Misclassification-Guaranteed Region $\mathcal{B}(x+\delta_1,k_1)$, while smaller size $\mathcal{B}(x+\delta_2,k_2)$ by $\delta_2$. \textbf{(b)}: Main Procedures of the PR-based UAP: (1) Attack initialization. (2) AE candidate selection. (3) $k$-driven optimization.
    }
    \label{Figure1}
  \end{center}
  \vskip -0.30in
\end{figure}

As illustrated in Fig.~\ref{Figure1}(a), two PGD runs yield perturbations $\delta_1$ and $\delta_2$ under an $\epsilon$ budget. Centered at the corresponding AEs $x+\delta$, the condition $\mathrm{PR}(x+\delta, k)=0$ implies that the neighborhood with radius $k$ (denoted as $\mathcal{B}(x+\delta,k)$ later) contains no label-preserving samples, i.e., all input points within it are AEs. Hence, a larger $k$ indicates a stronger attack, inducing misclassification over a larger neighborhood. However, in such an approach, $k$ is a post-hoc metric and does not participate in the perturbation generation process.

Now we present the rationale for the proposed PR-based UAP attack by providing an intuitive explanation of the compatibility between the PR-driven optimization and UAP generation. Specifically, using $\delta_1$ and $\delta_2$ in Fig.~\ref{Figure1}(a) as representative examples, a UAP targets misclassification for most data samples $x$ in the distribution $\mathcal{X}$ with a shared perturbation~\cite{moosavi2017universal}. Let $\delta_n$ ($n\in\{1,2\}$) be such a shared perturbation. Applying $\delta_n$ to any input induces a \textbf{distribution-level translation} $x \mapsto x+\delta_n$ (shaded regions in Fig.~\ref{Figure1}(a)), which shifts the dashed-circle neighborhood around $x$ to the solid-circle neighborhood around $x+\delta_n$ (follow the two arrows, respectively). In this context, PR also provides a \textbf{distribution-aware} view of adversarial impact, naturally aligning with the distribution-level target and translation mapping of UAPs. By Eq.~\ref{equation:PR_k}, the notion of $\mathrm{PR}(x+\delta,k)=0$ implies $\forall z\in\mathcal{B}(x+\delta,k),\\\, f_\theta(z)\neq y$, i.e., all input points within $\mathcal{B}(x+\delta,k)$ are AEs, which yields a converse implication: 
\begin{equation}
\label{eq:miscls_guarantee}
\begin{aligned}
&(\forall z \in \mathcal{B}(x+\delta_n,k_n),\ f_\theta(z)\neq y)
\wedge\ (\mathcal{B}(x,k_n)+\delta_n \subseteq \mathcal{B}(x+\delta_n,k_n))
\\
&\qquad\qquad\qquad\Rightarrow\ \forall x' \in \mathcal{B}(x,k_n),\ f_\theta(x'+\delta_n)\neq y .
\end{aligned}
\end{equation}
Given the UAP mapping from $\mathcal{B}(x,k_n)$ to $\mathcal{B}(x+\delta_n,k_n)$, any point $x'\in\mathcal{B}(x,k_n)$ will satisfy $f_\theta(x'+\delta_n)\neq y$. Therefore, $\delta_n$ induces a \textbf{misclassification-guaranteed} region $\mathcal{B}(x,k_n)$ around $x$ (dashed circles in Fig.~\ref{Figure1}(a)). A larger $k$ yields a broader affected neighborhood around $x$, as shown by the larger dashed circle for $\delta_1$ compared with the circle for $\delta_2$, thereby a larger portion of the input space $\mathcal{X}$ will be misclassified, which directly aligns with the objective of UAP generation. Thus, it is reasonable and intuitive to incorporate PR as an explicit objective in the optimization process for UAP generation.

\subsection{The proposed PR-based UAP attack}
\label{sec:The proposed PR-based UAP attack}
In this section, we detail the PR-based UAP attack, which reformulates the PR-derived variable $k$ as an explicit optimization objective and integrates it into the UAP update. This turns the post-hoc discrete selection of maximal $k$ into a direct continuous optimization problem, yielding a more principled UAP generation process. Our method comprises two algorithms: a UAP generation framework (details shown in Alg.~\ref{alg:uap} in Appx.~\ref{appendix:B}) and PR-based gradient computation (Alg.~\ref{alg:k.gred}). Specifically, Alg.~\ref{alg:k.gred} computes the PR-based update gradient and forwards it to Alg.~\ref{alg:uap} in Appx.~\ref{appendix:B}, which iteratively updates the perturbation to obtain the final PR-based UAP.

Alg.~\ref{alg:uap} in Appx.~\ref{appendix:B} displays the UAP generation framework.
In line~1, we sample a seed set $Seedset \subset Tr$ and initialize $uap$ $=\mathbf{0}\in\mathbb{R}^{d}$ with $d=\dim(x)$. During UAP generation, $Seedset$ is shuffled at the beginning of each iteration in line~4. The target classifier $C$ denotes the pre-trained DRL-based agent. For each $x \in Seedset$, the predictions $C(x)$ and $C(x+uap)$ are computed at line~5. If the predicted labels are identical (line~6), the current $uap$ fails to fool $C$ on $x$, and we then perform one update step by generating an update perturbation $ptb$. Specifically, we compute the gradient $g$ via the PR-based gradient calculation in Alg.~\ref{alg:k.gred} at line~7. Using \,$g$ \ as the update direction, \,we apply an element-wise mask to\, restrict \,modifications \,to \,MF\, only, \,and \,form \,$ptb$\, \,by \,an \,$L_2$-normalized\, FGM-style step with magnitude $\epsilon$~\cite{goodfellow2014explaining}. The accumulated $uap$ is then projected onto the $L_2$ ball to maintain the perturbation budget. The loop continues by updating the iteration number $iter\_num$ until reaching the max iteration $max\_iter$, after which the final $uap$ is returned (line~12).
\vskip -0.10in
\begin{algorithm}[H]
\caption{UAP Direction via the Gradient of $k^*$}
\label{alg:k.gred}
\textbf{Input:}  target model $C$, data instance $x$ with label $y$, perturbation budget $\epsilon$, PGD attack number $N$, PGD iteration step $n_{\mathrm{step}}$, PGD step size $\alpha$, SGD max iteration $SGD\_num$\\
\textbf{Output:} the gradient of $k^*$ with respect to $x$: $\nabla_x k^*$

\begin{algorithmic}[1]
\STATE Initialize $\mathcal{X}_{\mathrm{adv}} \gets [\;]$, $\mathcal{D} \gets [\;]$
\STATE $\mathcal{X}_{\mathrm{adv}}[0] \gets \mathrm{PGD}(x, y, n_{\mathrm{step}}, \alpha, \epsilon, \ell_2)$
\WHILE{$\mathrm{len}(\mathcal{X}_{\mathrm{adv}}) < N$}
    \STATE $z \sim \mathcal{N}(0, I), \quad r \sim \mathrm{Uniform}(0, \epsilon)$, \quad
    $x_{\mathrm{init}} \leftarrow x + r \cdot \frac{z}{\lVert z \rVert_2}$
    \STATE $\epsilon^{\prime} \leftarrow \mathrm{Unif}(0.98*\epsilon,\epsilon)$, $\alpha^{\prime} \leftarrow \mathrm{Unif}(0.9*\alpha,\,1.1*\alpha)$, $n^{\prime}_{\mathrm{step}} \leftarrow \mathrm{Unif}(0.9\,*n_{\mathrm{step}},\,1.1 \,*n_{\mathrm{step}})$
    \STATE Append $\mathrm{PGD}(x_{\mathrm{init}}, y, \epsilon^{\prime}, \alpha^{\prime}, n^{\prime}_{\mathrm{step}})$ to $\mathcal{X}_{\mathrm{adv}}$
\ENDWHILE
\FOR{ each $x_{\mathrm{adv}} \in \mathcal{X}_{\mathrm{adv}}$ }
    \STATE $x^{\prime} \leftarrow x_{\mathrm{adv}}$, $i \leftarrow 0$
    \WHILE {$C(x') \neq y \;\land\; i < SGD\_num$}
        \STATE $x' \leftarrow x' - \alpha \,
        \frac{\nabla_{x'} \mathcal{L}\big(C_{\mathrm{logit}}(x'), y\big)}
        {\left\lVert \nabla_{x'} \mathcal{L}\big(C_{\mathrm{logit}}(x'), y\big) \right\rVert_2}$, \,$i \,\leftarrow \,i + 1$
    \ENDWHILE
    \STATE Append $\lVert x_{\mathrm{adv}} - x' \rVert_2$ to $\mathcal{D}$
\ENDFOR
\STATE $k^* \leftarrow \max \mathcal{D}$
\STATE \textbf{return} $\nabla_x k^*$
\end{algorithmic}
\end{algorithm}
\vskip -0.2in
Next, we detail Alg.~\ref{alg:k.gred}, which computes the PR-driven update gradient for our PR-based UAP. As shown in Fig.~\ref{Figure1}(b), it consists of three main steps. Concretely, aiming to maximize the radius $k$ in Eq.~\ref{equation:PR_k}, we first employ the candidate-search method proposed by Zhang et al.~\cite{zhangyi2025adversarial} to identify the AE corresponding to local maximum loss in Steps~(1)--(2):

\noindent\textbf{(1) Attack initialization:} Multiple random PGD attacks are launched on input $x$ to obtain a set of AE candidates $\mathcal{X}_{\mathrm{adv}}$.\\
\noindent\textbf{(2) AE candidate selection:} For each AE candidate, the distance from $x_{\mathrm{adv}}$ to the decision boundary is estimated and treated as the corresponding $k$ in Eq.~\ref{equation:PR_k}. The candidate with the largest $k$ (denoted $k^*$) is selected as the starting point for the subsequent step.



The key difference from Zhang et al.~\cite{zhangyi2025adversarial} lies in Step~(3). The discrete selection of $k^*$ may be constrained by limited search resolution and finite exploration steps. Therefore, motivated by our rationale, selecting the largest value in $\mathcal{D}$ as $k^*$ chooses the candidate that causes the largest \textbf{misclassification-guaranteed} region $\mathcal{B}(x,k^*)$, inducing the most promising local region for UAP generation. Consequently, we formulate $k^*$ as an explicit optimization objective and adopt a continuous $k$-driven optimization scheme, enabling more effective and principled radius maximization.

 \noindent\textbf{(3) $k$-driven optimization:} The gradient of $k^*$ with respect to $x$, \,$\nabla_x k^*$\, is calculated and served as\\ the update gradient $g$ in Alg.~\ref{alg:uap}       (Eq.~\ref{eq:grad_kstar_simple}). By perturbing $x$ along the direction of $g$ to maximize the radius $k^*$, a more effective perturbation can be obtained, corresponding to $\delta^{\prime}, k^{\prime}$ and $x_{\mathrm{adv}}^{\prime}$ in Fig.~\ref{Figure1}(a).
\vskip -0.20in
\begin{gather}
\nabla_x k^*(x)=\nabla_x k\bigl(x,\delta^*\bigr),
\label{eq:grad_kstar_simple}\\
\text{where }\delta^*\triangleq \arg\max_{\delta}\,k(x,\delta)
\ \text{s.t. }\delta\in\mathrm{PGD}(x),\ \|\delta\|\le\epsilon .
\notag
\end{gather}
Specifically, Alg.~\ref{alg:k.gred} illustrates the computation of $\nabla_x k^*$. After initialization, multiple PGD attacks from random starting points $x_{\mathrm{init}}$ are launched (line~4), where $x_{\mathrm{init}}$ is sampled by $z\sim\mathcal{N}(0,I)$, $r\sim\mathrm{Uniform}(0,\epsilon)$, and $x_{\mathrm{init}} = x + r\cdot\frac{z}{\|z\|_2}$. We further randomize the attack hyperparameters $(\epsilon',\alpha',n^{\prime}_{\mathrm{step}})$ within predefined ranges to generate an AE set $\mathcal{X}_{\mathrm{adv}}$ (lines~5-6). For each AE candidate $x_{\mathrm{adv}}$, we iteratively update the current point $x'$ via a gradient descent-based method until it becomes non-adversarial or the maximum number of steps $SGD\_num$ is reached (lines~10--12). This procedure yields a new point $x'$ that can be regarded as an approximate “projection” of $x_{\mathrm{adv}}$ onto the decision boundary. At line~13, we compute $\lVert x_{\mathrm{adv}} - x' \rVert_2$ as the radius $k$ and store these $k$ values in a distance set $\mathcal{D}$, from which we take the maximum as $k^*$ (line~15). Thus, we compute $\nabla_x k^*$ (obtained by automatic differentiation since $k^*$ is computed without hard clipping or $\arg\max$) and return it as the update gradient $g$ to Alg.~\ref{alg:uap}. Then $x$ is perturbed along the direction of $\nabla_x k^*$, with $k^*$ serving as the optimization objective to be maximized, ultimately generating the proposed PR-based UAP.

\subsection{PX-UAP attack}
This section introduces the PX-UAP attack, which leverages XAI to guide perturbation shaping and further enhance attack effectiveness. Before presenting the technical details, we first provide the rationale behind PX-UAP. XAI comprises techniques that improve the interpretability of ML/DL models by providing human-understandable explanations, typically in the form of feature-level attributions that identify decision-critical inputs for a given prediction~\cite{doshi2017towards,guidotti2018survey}. In adversarial settings, such attributions offer a principled mechanism for prioritizing the perturbations under a limited adversarial budget, which is especially relevant for IDS where realistic domain constraints restrict arbitrary feature modifications~\cite{zhang2025explainable,marino2018adversarial}. 

Under these constraints, XAI-derived signals enable the identification of influential and permissible features, thereby facilitating more effective UAPs. Prior studies show that attribution-guided pertur-\\bations facilitate per-instance AE generation in IDS~\cite{marino2018adversarial, okada2025xai}. However, these efforts do not consider realistic network domain constraints and have not explored XAI's role in input-agnostic UAP generation. Motivated by this observation, we incorporate XAI to guide the proposed PR-based UAP, resulting in the PX-UAP attack, representing the first XAI-guided universal adversarial attack in IDS.

Now we introduce PX-UAP (Alg.~\ref{alg:XAI} in Appx.~\ref{appendix:B}), which calculates the feature-level importance scores and selects effective perturbations based on detection performance. Specifically, the PR-based UAP $PRuap$ is first recalculated for realistic constraints and projected onto the $L_2$ budget $\epsilon$ (line~1). We use the predictions of realistic AEs $(Tr + PRuap)$ and clean samples $Tr$ to compute the false negative rate (FNR) as a reference performance metric for later comparison (line~2). A higher FNR indicates that more malicious samples are misclassified as benign, aligning with the adversary’s objective in this work. At line~3, a batch $\mathcal{B}$ with $B$ instances is sampled from the training set $Tr$ to estimate feature-level importance. In this research, the \textit{benign} class is used as the target class, and Integrated Gradients~\cite{sundararajan2017axiomatic} as $\operatorname{XAImtd}$ to compute feature attribution scores, as shown in Eq.~\ref{eq:IG_and_batch}:
\vskip -0.20in
\begin{align}
\mathrm{IG}(x; x')
&\triangleq (x\!-\!x')\odot \textstyle\int\nolimits_{0}^{1} \nabla_x\, C_{\mathrm{logit}}\!(x'\!+\!\alpha(x\!-\!x'))\,\mathrm{d}\alpha.
\label{eq:IG_and_batch}
\end{align}
IG is well suited to our white-box threat model and the case of tabular data, as it provides gradient-based attributions with low computational overhead~\cite{hosain2024explainable}. In practice, we use the discrete $m$-step approximation in Eq.~\ref{eq:IG_and_batch_m}. The attribution scores for each $x\in\mathcal{B}$ are averaged to obtain an aggregated importance vector $imp$ (lines~4--5).

\vskip -0.20in
\begin{align}
\widehat{\mathrm{IG}}(x; x')
&\triangleq (x\!-\!x')\odot \frac{1}{m}\textstyle\sum\limits_{t=1}^{m}
\nabla_x\, C_{\mathrm{logit}}\!(x'\!+\!\tfrac{t}{m}(x\!-\!x')).
\label{eq:IG_and_batch_m}
\end{align}
\vskip -0.1in

 In line~6, we normalize $imp$ by $\lVert imp\rVert_{\infty}$, scaling its values to a standard range $[-1,1]$. This mitigates batch-sampling variability in estimating $imp$, ensuring a standardized XAI-based weighting and a fair comparison across attacks. We then obtain $PXuap$ by XAI-weighting $PRuap$ as in Eq.~\ref{eq:pxuap_weighting}:

\vskip -0.1in
\begin{equation}
\label{eq:pxuap_weighting}
\begin{aligned}
imp \leftarrow \mathrm{mask}\odot \textstyle\frac{imp}{\lVert imp\rVert_\infty},\; imp\in[-1,1]^d,
u &\leftarrow \mathrm{PRuap}\odot \mathrm{Softmax}(imp),\quad
\mathrm{PXuap}\leftarrow \epsilon \cdot \textstyle\frac{u}{\lVert u\rVert_2}.
\end{aligned}
\end{equation}

Specifically, taking $imp$ as input, a feature mask is applied to restrict weighting only on the top-$n$ most important features. We then transform $imp$ via $\operatorname{Softmax}(\cdot)$ to obtain feature-wise scaling coefficients, which nonlinearly map relative differences in importance scores and concentrate the limited perturbation budget on the most influential features. For example, $imp=[1,-1,0]$ yields $\operatorname{Softmax}(imp)\approx[0.66,0.09,0.24]$. After the projection and recalculation, the FNR of $PXuap$ is computed at line~7. Due to the stochasticity introduced by batch sampling, we compare the performance of $PXuap$ with $PRuap$ and select the one with a higher FNR as the final PX-UAP.

\subsection{Theoretical analysis}
\label{sec:theoretical_analysis}

This subsection provides a rigorous analysis of the PX-UAP design in
Eq.~\ref{eq:pxuap_weighting} by (i) formalizing the targeted universal evasion
objective under feasibility constraints, (ii) establishing IG
as a principled feature-priority signal, (iii) showing that Softmax weighting
arises as an entropy-regularized allocation rule, and (iv) giving a
first-order improvement guarantee for the resulting weighted perturbations.

\textbf{Formal objective: targeted universal evasion under feasibility.}
The adversary targets \textit{malicious}$\to$\textit{benign} evasion in
white-box setting. Let $\mathcal{X}_1 \subseteq Tr$ denote the malicious subset
($x$ with ground-truth $y=1$), and let the target class be
$y_t = 0$ (benign).
We denote by $\widetilde{u} \triangleq \operatorname{recalculate}(u,\epsilon,\ell_2)$
the feasibility-enforced perturbation returned by the realistic constraints in Alg.~\ref{alg:XAI}.

The targeted universal evasion objective can be written as
\begin{equation}
\label{eq:universal_evasion_obj_aligned}
\begin{aligned}
    & \max_{u \in \mathbb{R}^d}\;
    \mathbb{E}_{x \sim \mathcal{X}_1}\Big[
    \mathbb{I}_{\{C(x+\widetilde{u}) = y_t\}}
    \Big] 
    & \text{s.t.}\;\; \|\widetilde{u}\|_2 \le \epsilon,\;\; \widetilde{u}\ \text{satisfies realistic constraints.}
\end{aligned}
\end{equation}

Eq.~\ref{eq:universal_evasion_obj_aligned} is aligned with the metric used in
Alg.~\ref{alg:XAI}: maximizing the probability of predicting benign on malicious inputs
corresponds to increasing FNR. Moreover, the PR-based construction (Sec.~\ref{sec:PR-rationale}) aims to enlarge the \emph{misclassification-guaranteed} neighborhood via the PR-derived radius $k$ (Eq.~\ref{equation:PR_k}). It increases the measure of inputs mapped into the adversarial region, consistent with the universal objective above.

\textbf{Integrated Gradients as a principled feature priority signal.}
PX-UAP uses IG-based feature attributions as $imp$ (Eq.~\ref{eq:IG_and_batch}).
The following formalizes why IG is appropriate for deciding which features
deserve more perturbation.

\begin{proposition}[Completeness of Integrated Gradients]
\label{prop:ig_completeness}
Let $F:\mathbb{R}^d\to\mathbb{R}$ be differentiable and $x,x' \in \mathbb{R}^d$. Define $\mathrm{IG}(x;x')$ as in Eq.~\ref{eq:IG_and_batch}.
Then IG satisfies the completeness identity $\sum_{i=1}^d \mathrm{IG}_i(x;x') \;=\; F(x) - F(x')$.
When $F(\cdot)=C_{\mathrm{logit}}(\cdot)$ for the target class $y_t$,
positive $\mathrm{IG}_i(x;x')$ identifies features whose increase along the path
from $x'$ to $x$ raises the target logit, and thus are natural candidates for
priority perturbation in targeted evasion.
\end{proposition}

In lines~5--6 in Alg.~\ref{alg:XAI}, we aggregate IG over $\mathcal{B}$ to obtain $imp$, so that (by Prop.~\ref{prop:ig_completeness})
$imp$ acts as an empirical estimate of which dimensions contribute
to increasing $C_{\mathrm{logit}}(\cdot)$ toward the target class $y_t$ across the data
distribution.

\textbf{Why Softmax weighting: an entropy-regularized allocation view.}
PX-UAP transforms $imp$ into nonnegative coefficients via Softmax and uses them to
reallocate the perturbation of $PRuap$ (Eq.~\ref{eq:pxuap_weighting}).
This can be derived as the unique solution of an entropy-regularized linear allocation.
Let $\Delta_{\mathrm{mask}}$ denote masked probability: $\Delta_{\mathrm{mask}}\triangleq \{w\in\mathbb{R}^d_{\ge 0}:\;\mathbf{1}^\top w = 1,\;\; ( \mathbf{1} \mathrm{mask})\odot w = \mathbf{0}\}$.

\begin{table*}[t]
  \caption{UAP baseline with loss functions for comparison with PR-based UAP.}
  \vskip -0.15in
  \label{tab:multi-loss}
  \begin{center}
    \begin{small}
      \begin{sc}
        \renewcommand{\arraystretch}{1.2} 

        \begin{tabular}{p{0.95cm}p{3.3cm}p{8.5cm}}
          \toprule
          \makecell{\textsc{\textbf{Index}}} &
          \makecell{\textsc{\textbf{Baseline UAP}}} &
          \makecell{\textsc{\textbf{Loss function to maximize}}} \\
          \midrule
          
          \makecell[c]{(1)} &
          \makecell[c]{\scriptsize\;\;\;\;PD\_mean\_UAP \cite{mopuri2017fast}} &
          \makecell[c]{
            $\log\!\left( \prod_{i=1}^{k} \mathrm{mean}(l_i+\mathrm{eps}) \right)$ \\
            $\text{where } l_i=\mathrm{activation}(x+\delta),\ \mathrm{eps}=1\times10^{-8}$
          } \\[5pt]

          \makecell[c]{(2)} &
          \makecell[c]{\scriptsize\hspace*{2.5mm}PD\_L2\_UAP \cite{mopuri2018generalizable}} &
          \makecell[c]{
            $\log\!\left( \prod_{i=1}^{k} \left\lVert l_i \right\rVert_2 + \mathrm{eps} \right)$ \\
            $\text{where } l_i=\mathrm{activation}(x+\delta),\ \mathrm{eps}=1\times10^{-8}$
          } \\[3pt]

          \makecell[c]{(3)} &
          \makecell[c]{\scriptsize\hspace*{2.5mm}COSSIM\_L3\_UAP \cite{ye2023fg}} &
          \makecell[c]{
            $-\mathrm{cossim}\!\left(\mathrm{activation}_{l_i}(x+\delta),\ \mathrm{activation}_{l_i}(x)\right)$,
            $\text{where } l_i=l_3$
          } \\[0.5pt]

          \makecell[c]{(4)} &
          \makecell[c]{\scriptsize\hspace*{2.5mm}COSSIM\_L4\_UAP \cite{ye2023fg}} &
          \makecell[c]{
            $-\mathrm{cossim}\!\left(\mathrm{activation}_{l_i}(x+\delta),\ \mathrm{activation}_{l_i}(x)\right)$,
            $\text{where } l_i=l_4$
          } \\[0.5pt]

          \makecell[c]{(5)} &
          \makecell[c]{\scriptsize\hspace*{3mm}PCC\_UAP \cite{zhang2025novel}} &
          \makecell[c]{
            $\mathrm{PCCsim}\!\left(\mathrm{activation}_{l_i}(x+\delta),\ \mathrm{activation}_{l_i}(\delta)\right)$,
            $\text{where } l_i=l_4$
          } \\[0.5pt]
          \bottomrule
        \end{tabular}
      \end{sc}
    \end{small}
  \end{center}
  \vskip -0.3in
\end{table*}

Define Shannon entropy $H(w)\triangleq -\sum_{i=1}^d w_i\log w_i$ for $w\in\Delta_{\mathrm{mask}}$~\cite{lesne2014shannon}.
Then:

\begin{lemma}[Softmax as the entropy-regularized maximizer]
\label{lem:softmax_entropy}
Fix a score vector $s\in\mathbb{R}^d$ and temperature $\tau>0$.
The optimization problem $\max_{w\in\Delta_{\mathrm{mask}}}\;\; \langle s,w\rangle + \tau H(w)$ has a unique solution given by
\begin{equation}
\label{eq:softmax_solution}
\begin{aligned}
    & w_i^\star = \frac{\exp(s_i/\tau)}{\sum_{j:\,\mathrm{mask}_j=1}\exp(s_j/\tau)}
    & \text{for all } i \text{ with }\mathrm{mask}_i=1, \quad w_i^\star=0\;\text{if }\mathrm{mask}_i=0,
\end{aligned}
\end{equation}
i.e., the Softmax distribution on the masked coordinates.
\end{lemma}

Lemma~\ref{lem:softmax_entropy} provides an interpretation of the Softmax
step in Eq.~\ref{eq:pxuap_weighting}: it allocates a unit budget across feasible features by trading off exploitation (large $\langle s,w\rangle$) and diversity/stability (large entropy). In PX-UAP we set $s \equiv imp$ (after normalization and masking), and use $\tau=1$, yielding the weighting in Eq.~\ref{eq:pxuap_weighting}. \textit{Due to space limits, the rest of the theoretical analysis is provided in Appx.~\ref{appendix:theory}.}

\section{Experiments}
\label{Chapter4:Experimental Result}
In this section, we describe the experimental setup and provide the experimental results to validate the effectiveness of our algorithm design. More results are provided in Appx.~\ref{Additional Experimental Result}.
\subsection{Experimental Settings}
\label{Experimental Settings}

In this study, we train a DQN agent~\cite{mnih2015human} on CICIDS2018 dataset~\cite{sharafaldin2018toward}, keeping \textit{benign} traffic as the \textit{benign} class and merging all remaining categories into the \textit{malicious} class. Then we use the agent's policy network as the IDS classifier $C$, which is a fully-connected MLP with four hidden layers (64 units each) and ReLU activations. $C$ outputs $Q(x,a)$ for $a\in\{0,1\}$ corresponding to the two classes, and predicts $\hat{y}=\arg\max_a Q(x,a)$. Each network flow is represented as a feature vector $x\in\mathbb{R}^{76}$, where each feature is normalized to $[0,1]$, with one-hot encoding for categorical features, and under-\\-sampling for class imbalance.  We evaluate the attack effectiveness in all experiments by FNR and accuracy (ACC), with the adversarial budget $\epsilon$ on the x-axis under the $L_2$ constraint. For allocating sufficient perturbation budget for effective attacks while preserving the \textit{imperceptible} property in the normalized feature space, we set $\epsilon \in [0.04,\,0.15]$. To mitigate potential bias from algorithmic \,ran-\\-domness, we report results averaged over 150 runs, ensuring a fair comparison across methods. The hyperparameter settings for all algorithms are provided in Appx.~\ref{appendix:c}.
\subsection{PR-based UAP result}
  \vskip -0.1in
\begin{figure}[H]
  \begin{center}    \centerline{\includegraphics[width=0.85\columnwidth]{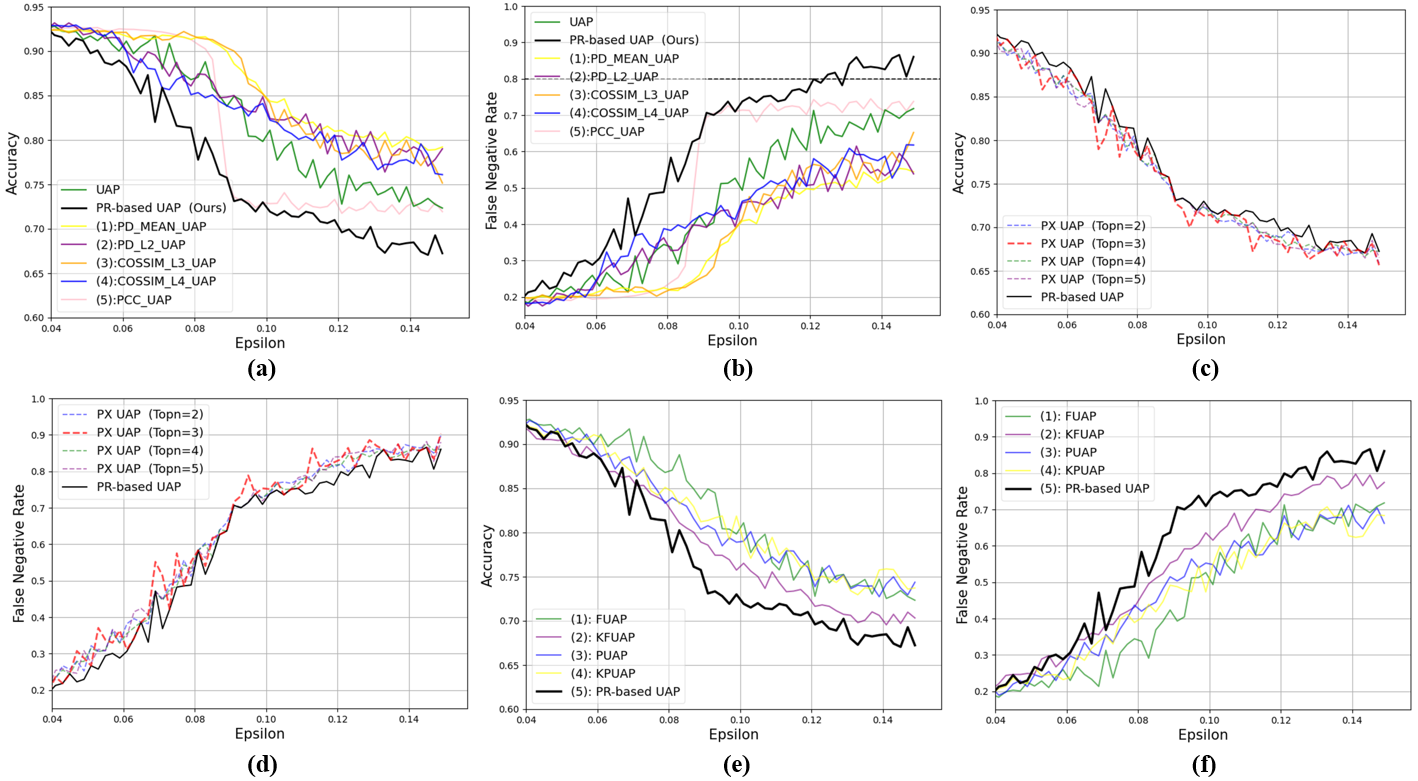}}
    \caption{ (a): False Negative Rate (FNR) and (b): Accuracy of the proposed PR-based UAP and other UAP baselines. (c): FNR and  (d): ACC of PX-UAP with different $\mathrm{Topn}$ values. (e): FNR and (f): ACC for ablation study.
    }
    \label{Figure3}
  \end{center}
  \vskip -0.35in
\end{figure}
We now report the results of the proposed PR-based UAP against several state-of-the-art UAP baselines, whose details are summarized in Tab.~\ref{tab:multi-loss}. As shown in Fig.~\ref{Figure3}(a) and (b), the PR-based UAP consistently outperforms all baselines across the evaluated $\epsilon$ range. Starting from $\epsilon=0.04$, all evaluated UAP methods exhibit comparable effectiveness, with $\mathrm{FNR}\approx0.2$ and $\mathrm{ACC}\approx0.93$. As $\epsilon$ increases to 0.09, all attacks become progressively more effective, while the PR-based UAP remains the top performer. Notably, PCC\_UAP (pink) rises sharply near $\epsilon=0.09$ and matches the proposed method at $\epsilon=0.09$, achieving $\mathrm{FNR}=0.72$ and $\mathrm{ACC}=0.73$. After this point, the proposed PR-based UAP continues to improve and stabilizes for $\epsilon \ge 0.13$. Overall, these results demonstrate that our PR-based UAP framework is effective and successfully surpasses other state-of-the-art UAPs.
\subsection{PX-UAP result}
Building upon the PR-based UAP, we report the results of PX-UAP in Fig.~\ref{Figure3}(c) and (d) with different $\mathrm{Topn}$ values, i.e., only the $\mathrm{Topn}$ most important features in $imp$ are used for XAI-based weighting in Alg.~\ref{alg:XAI}.
During the early “cold-start” stage ($\epsilon\!\in\![0.04, 0.07]$), all PX-UAP variants consistently strengthen the attack, with an $\approx$10\% improvement in effectiveness (higher FNR and lower ACC) over the PR-based baseline. In the mid-range ($\epsilon\in[0.07,0.10]$), the XAI-induced benefit is comparatively limited.  \ \ Only
$\mathrm{Topn}\!=\!3$
(red) achieves a $\approx$5\% improvement around $\epsilon\approx0.075$. Later, XAI guidance further improves the PR-based UAP, even though it already outperforms state-of-the-art baselines in $\epsilon\in[0.10,0.15]$. All four PX-UAP variants provide additional gains in this “stable” range, with $\mathrm{Topn}\!=\!3$ achieving the best overall performance. Overall, the results show that incorporating XAI enables PX-UAP to outperform the proposed PR-based UAP, with particularly pronounced gains at small budgets and in the high-$\epsilon$ range where performance saturates. In these cases, XAI helps allocate the limited perturbation budget to influential features, thereby improving attack effectiveness.
\subsection{Ablation Studies}
As described in Sec.~\ref{sec:The proposed PR-based UAP attack}, Alg.~\ref{alg:k.gred} comprises three key components: (i) PGD-based attack initialization, (ii) $k$-based selection of AE candidates, and (iii) $k^*$-driven optimization. To verify that the observed effectiveness stems from the combination of the components rather than any single factor, we conduct ablation studies by comparing the proposed PR-based UAP with several ablated variants in which specific components are removed or modified. Details and explanations of the variants are summarized in Tab.~\ref{PR UAP candiate} in Appx.~\ref{appendix:ablation study}. The results of the ablation study are reported in Fig.~\ref{Figure3}(e) and (f).

For $\epsilon<0.06$, all candidates exhibit similar performance, with $\mathrm{FNR}\approx0.28$ and $\mathrm{ACC}\approx0.90$. Then, PR-based UAP gradually outperforms the others across the remaining $\epsilon$ range. Among the variants, \textbf{AE selection} improves effectiveness under FGM initialization, as evidenced by FUAP (1) versus KFUAP (2), but limited benefit under PGD initialization (PUAP and KPUAP; (3) and (4)). All var-\\iants indicate that changing \textbf{Attack initialization} and \textbf{AE selection} has a limited impact on overall attack effectiveness. Taking the best-performing PR-based UAP and the basic FUAP as the references, the ablation study indicates that combining all three components jointly in Alg.~3 yields a substantial performance improvement, with the \textbf{$k^*$-driven optimization} playing a critical role in this performance gain, which further validates the effectiveness of our proposed rationale and algorithm design.
\subsection{Additional Results}
Beyond the results above, we conduct extensive additional experiments for a comprehensive evaluation of the proposed attack method, detailed results and analysis are shown in Appx.~\ref{Additional Experimental Result}.  In summary:\\
\hspace*{1em}$\bullet$ We report the F1 scores of the above three experiments, providing a more balanced and comprehensive metric to evaluate the effectiveness of the proposed PR-based UAP and PX-UAP.\\
\hspace*{1em}$\bullet$ We apply the proposed attack methods to two other dataset and further use different DRL agents and network architectures as target classifiers on CICIDS2018. The results demonstrate the generalizability of the proposed attack across datasets, DRL agents, and model architectures.\\
\hspace*{1em}$\bullet$ We further evaluate the proposed attack methods under a black-box threat model by generating UAPs on a surrogate model and applying them to different target models. The results demonstrate that the proposed attacks exhibit strong cross-model transferability under the black-box setting.\\
\hspace*{1em}$\bullet$ We conduct an ablation study on $\epsilon$-steps to verify that the significant performance gain of our proposed PR-based UAP stems from the attack design itself rather than favorable step-size selection. The results show that PR-based UAP consistently outperforms the baselines under both the fixed unified $\epsilon$-step setting and the setting where $\epsilon$-steps are tuned for each method.
\section{Conclusion}
\label{Chapter5:Conclusion}
In this work, for the first time, we integrate the concept of PR into the UAP generation process and propose a novel UAP attack against DRL–based IDS. We further enhance attack performance by advancing XAI techniques to guide the perturbation shaping for the proposed PR-based UAP framework under realistic domain constraints, with providing a rigorous theoretical justification. Extensive experiments demonstrating that PX-UAP consistently outperforms state-of-the-art UAP methods for intrusion detection in terms of attack effectiveness.

\bibliography{example_paper}


\appendix

\section{Supplementary Materials for Theoretical Analysis}
\label{appendix:theory}

We provide (iv) a first-order improvement guarantee for the resulting weighted perturbations under standard local assumptions. The previous theoretical analysis is placed in Sec.~\ref{sec:theoretical_analysis}.

\textbf{First-order improvement guarantee under weighted perturbations.}
We now connect the entropy-allocation view to a \emph{first-order} targeted-evasion surrogate.
Let $v \triangleq PRuap$ be the feasibility-enforced PR-based UAP, and define the
family of reweighted perturbations
\begin{equation}
\label{eq:weighted_family}
u(w) \triangleq v \odot w,
\qquad
\delta(w) \triangleq \epsilon \cdot \frac{u(w)}{\|u(w)\|_2},
\qquad w\in \Delta_{\mathrm{mask}}.
\end{equation}
Consider the targeted logit function $F(x)\triangleq C_{\mathrm{logit}}(x)$ for $y_t$ and
its local first-order Taylor approximation:
$F(x+\delta)\approx F(x)+\nabla_x F(x)^\top \delta$ for small $\|\delta\|_2$~\cite{jentzen2011taylor}.
Averaging over the batch $\mathcal{B}$ in Alg.~\ref{alg:XAI}, let
$\bar{g}\triangleq \frac{1}{|\mathcal{B}|}\sum_{x\in\mathcal{B}}\nabla_x F(x)$.
Ignoring the (direction-preserving) normalization in Eq.\ref{eq:weighted_family} yields the
standard linear surrogate gain:
\begin{equation}
\label{eq:linear_gain}
\mathcal{G}(w)
\triangleq
\left\langle \bar{g},\, u(w)\right\rangle
=
\sum_{i=1}^d \underbrace{\big(\bar{g}_i v_i\big)}_{s_i}\, w_i
=
\langle s,w\rangle,
\end{equation}
where $s \triangleq \bar{g}\odot v$ is a per-feature score that combines (i) how much
increasing feature $i$ raises the target logit via $\bar{g}_i$, and (ii) the PR-based
universal direction already discovered under PR optimization via $v_i$.

\begin{corollary}[Connection to PX-UAP with IG scores]
\label{cor:pxuap_connection}
Suppose $imp$ (Eq.~\ref{eq:IG_and_batch}) is used as a proxy score for $s$ in
\ref{eq:linear_gain}, which is justified by IG completeness
(Prop.~\ref{prop:ig_completeness}) since IG decomposes the target-logit difference across
features along a path and thus preserves feature ranking for targeted logit increase.
Then the PX-UAP weighting in Eq.~\ref{eq:pxuap_weighting} implements the maximizer of an
entropy-regularized first-order setting over $\Delta_{\mathrm{mask}}$, and reallocating the fixed $\ell_2$ budget $\epsilon$ toward the most influential feasible features while maintaining stable (non-degenerate) allocation through the entropy term.
\end{corollary}

Overall, this analysis explains why $PXuap$ can outperform $PRuap$ under feasibility constraints: $PRuap$ identifies a strong universal direction via PR/$k$ optimization, and the IG$\to$Softmax step then performs a principled budget reallocation toward decision-critical feasible dimensions. It can improve the expected targeted-evasion gain under a fixed $\ell_2$ budget.

\section{Addition Algorithims}
\label{appendix:B}
In this section, we show the two addition algorithms that not shown in the content of main paper, inclduing the UAP generation framework in Alg.~\ref{alg:uap} and the PX-UAP attack generation in Alg.~\ref{alg:XAI}.
\begin{algorithm}[H]
\caption{UAP Generation for Classifier $C$}
\label{alg:uap}
\textbf{Input:} Trainset $Tr$, Classifier $C$, Modified feature group mask $mask$, seedset size $size$, max iteration number $max\_iter$, perturbation budget $\epsilon$ 

\textbf{Output:} Universal adversarial perturbation $uap$

\begin{algorithmic}[1]
\STATE $Seedset \gets \mathrm{Sample}(Tr,\; size \cdot |Tr|)$, $uap \gets \mathbf{0}, iter\_num \gets 0$
\WHILE{$iter\_num < max\_iter$}
    \STATE Randomly shuffle $Seedset$
    \FOR{each $x \in Seedset$}
        \STATE $L_1, L_2 \gets C(x),\, C(x + uap)$
        \IF{$L_1 = L_2$}
            \STATE $g \gets \text{gradient from Alg.~\ref{alg:k.gred}}$, \:$ptb \gets \epsilon \cdot \dfrac{mask \odot g}{\|mask \odot g\|_2}$, \:$uap \gets \mathrm{Project}_{\ell_2}\!\left(uap + ptb\right)$
        \ENDIF
    \ENDFOR
    \STATE $iter\_num \gets iter\_num + 1$
\ENDWHILE
\STATE \textbf{return} $uap$
\end{algorithmic}
\end{algorithm}
\vskip -0.2in

\begin{algorithm}[H]
\caption{PX-UAP attack}
\label{alg:XAI}
\textbf{Input:}  Trainset $Tr$, target model $C$, epsilon $\epsilon$, batch size $B$, XAI feature selection mask $\mathrm{mask}$ \\
\textbf{Output:} XAI enhanced UAP \ $PXuap$

\begin{algorithmic}[1]
\STATE $ uap \leftarrow \operatorname{PR-based\ UAP}(C, Tr, \epsilon)$,\quad $ PRuap \leftarrow \operatorname{recalculate}( uap, \epsilon, \ell_2)$
\STATE $\mathrm{FNR} \leftarrow \operatorname{ComputeFNR}(C(Tr),\, C(Tr + PRuap))$
\STATE Sample a batch $\mathcal{B} \subset Tr$ with $|\mathcal{B}| = B$
\STATE $imp_{\text{batch}} \leftarrow \operatorname{XAImtd}(\mathcal{B},\, target\_class)$
      \hfill $\triangleright$ $B \times d$
\STATE $imp \leftarrow \frac{1}{|\mathcal{B}|}\sum_{x \in \mathcal{B}} imp_{\text{batch}}(x)$
      \hfill $\triangleright$ $d$
\STATE Compute\ $PXuap$ \ according to Eq.~\ref{eq:pxuap_weighting}
\STATE $PXuap \leftarrow \operatorname{recalculate}( PXuap,\epsilon, \ell_2)$,\, $\mathrm{FNR^{\prime}} \leftarrow \operatorname{ComputeFNR}(C(Tr), C(Tr + PXuap))$
\IF{$\mathrm{FNR^{\prime}} < \mathrm{FNR}$}
    \STATE $PXuap \leftarrow PRuap$
\ENDIF
\STATE \textbf{return} $PXuap$
\end{algorithmic}
\end{algorithm}
\vskip -0.20in

\section{Network Realistic Domain Constraints}
\label{appendix:C}

In this section, we present the feature categorization for the CICIDS2018 dataset adopted in this study, together with brief descriptions and recalculation formulations, as shown in Tab.~\ref{tab:feature_groups_2}. We partition the input features into three groups: \textit{Modified Features (MF)}, \textit{Related Features (RF)}, and \textit{Unmodified Features (UF)}~\cite{zhang2025novel}. 

Specifically, for any input flow feature vector $x \in \mathcal{X}$ (where $x_i$ denotes the $i$-th feature of $x$), the adversarial perturbation $\delta$ is applied only to the \textit{Modified Features} (MF) first, with the perturbed value $(x+\delta)_i$ must remain within its valid range (e.g., protocol compliance and admissible feature bounds~\cite{merzouk2022investigating}), as formalized in Eq.~\ref{equation:1}:
\begin{equation}
\label{equation:1}
\forall x \in \mathcal{X},\quad \forall i \in MF,\quad (x + \delta)_i \in \mathcal{R}_i(\mathcal{X})
\end{equation}

Regarding to RF, whenever the adversary modifies any feature in $MF$ (denoted as the $i$-th feature), the corresponding features in $RF$  (denoted as the $j$-th feature) must be recalculated to preserve inter-feature mathematical dependencies, shown in Eq.~\ref{equation:2}:
\begin{equation}
\label{equation:2}
\forall x \in \mathcal{X},\quad \forall j \in RF,\quad (x + \delta)_j = \text{Recalculate} (x + \delta)_i\ 
\end{equation}

The remaining features are assigned to \textit{Unmodified Features} (UF) and remain fixed throughout the attack, since their values cannot be altered via feasible manipulation actions. After these steps, the resulting perturbation is projected onto the $L_2$ ball of radius $\epsilon$ to enforce the prescribed attack budget.

\begingroup
\begin{table*}[ht]
\caption{Feature categorization, descriptions, and formulations}
\label{tab:feature_groups_2}
\centering

{ \renewcommand{\arraystretch}{1.5}
\begin{tabular}{>{\centering\arraybackslash\footnotesize}m{2.35cm} 
                >{\centering\arraybackslash\footnotesize}m{2.5cm} 
                >{\centering\arraybackslash\footnotesize}m{5.6cm} 
                >{\centering\arraybackslash\footnotesize}m{2cm}}
\toprule
\makecell{\textsc{\textbf{Feature Name}}} & 
\makecell{\textsc{\textbf{Description}}} & 
\makecell{\textsc{\textbf{Recalculate Formulation}}} & 
\makecell{\textsc{\textbf{Group}}} \\
\midrule

Tot Fwd Pkts & Total packets in the forward direction & – & \multirow{5}{*}{Modified Features} \\ 
Tot Bwd Pkts & Total packets in the backward direction & – & \\ 
TotLen Fwd Pkts & Total size of packet in forward direction & – & \\ 
TotLen Bwd Pkts & Total size of packet in backward direction & – & \\ 
Flow Duration & Duration of the flow in Microsecond & – & \\ 
\midrule

Fwd Pkts/s & Number of forward packets per second & 
${\text{Tot Fwd Pkts} \times 10^6}\;/\;{\text{Flow Duration}}$& 
\\ 

Bwd Pkts/s & Number of backward packets per second & 
${\text{Tot Bwd Pkts} \times 10^6}\;/\;{\text{Flow Duration}}$ & 
\\ 

Flow Pkts/s & Number of flow packets per second & 
{$\displaystyle \text{Fwd Pkts/s} + \text{Bwd Pkts/s}$} & 
\\ 

Flow Byts/s & Number of flow bytes per second & 
\rule{0pt}{17pt}\raisebox{1ex}{$\displaystyle \frac{(\text{TotLen Fwd Pkts} + \text{TotLen Bwd Pkts}) \times 10^6}{\text{Flow Duration}}$} & \raisebox{-1\baselineskip}{Related Features}
\\ 

Pkt Size Avg & Average size of packet & 
\rule{0pt}{17pt}\raisebox{1ex}{$\displaystyle \frac{\text{TotLen Fwd Pkts} + \text{TotLen Bwd Pkts}}{\text{Tot Fwd Pkts} + \text{Tot Bwd Pkts}}$} & 
\\ 

Fwd Seg Size Avg & Average size in the forward direction & 
$\text{TotLen Fwd Pkts}\;/\;{\text{Tot Fwd Pkts}}$ & 
\\ 

Bwd Seg Size Avg & Average size in the backward direction & 
$ {\text{TotLen Bwd Pkts}}\;/\;{\text{Tot Bwd Pkts}}$ & 
\\ 

Down/Up Ratio & Download and upload ratio & 
{$\text{int} ( {\text{Tot Bwd Pkts}}\;/\;{\text{Tot Fwd Pkts}} )$} & 
\\ 
\midrule

Other Features & Remaining features not in above groups & – & Unmodified Features \\ 
\bottomrule
\end{tabular}
}
\end{table*}
\endgroup

\clearpage
\section{Hyperparameter Setting for Algorithms}
\label{appendix:c}
In this section, we report the hyperparameter settings for three algorithms in the main paper, shown in Tab.~\ref{tab:hyperparam_setting}. Specifically, to balance computational efficiency with sufficient coverage of the training distribution, we generate the UAP using a small portion of the training set and set the seedset size to $1\times 10^{-4}$ (corresponding to 600 network data samples as the seedset for UAP generation). For the same reason, we set the XAI batch size $B$=10,000 to improve sample coverage for more representative XAI estimation, while keeping the computational overhead manageable. All the attack methods and their variants shown in Sec.~\ref{Chapter4:Experimental Result} are evaluated under identical iteration settings (e.g. $max\_iter$, $N$, $n_{\mathrm{step}}$, $SGD\_num$, $\alpha$).

\begin{table*}[h]
\caption{Hyperparameter Setting for algorithms}
\label{tab:hyperparam_setting}
\centering
\small
\setlength{\tabcolsep}{6pt}
\begin{tabular}{
>{\centering\arraybackslash}p{0.20\textwidth}
>{\centering\arraybackslash}p{0.28\textwidth}
>{\centering\arraybackslash}p{0.20\textwidth}
}
\toprule
\textsc{\textbf{Feature Name}} & \textsc{\textbf{Feature Description}} & \textsc{\textbf{Feature Value}} \\
\midrule
$\epsilon$ & Adversarial Budget & $[0.04,\,0.15]$\\
$size$ & Seedset size & $1\times 10^{-4}$ \\
$max\_iter$ & Max Iteration Number & 20 \\
$N$& PGD Attack Number & 10 \\
$n_{\mathrm{step}}$& PGD Iteration Step & 50\\
$\alpha$ & PGD Step Size  & $\epsilon / n_{\mathrm{step}}$\\
$SGD\_num$  & SGD Max Iteration Number & 50 \\
$B$ & XAI Batch Size & 10000\\
$m$ & IG Iteration Step & 500\\

\bottomrule
\end{tabular}
\end{table*}

\section{Variants detail for Ablation Study}
\label{appendix:ablation study}
 To verify that the observed attack effectiveness stems from the combination of the components rather than any single factor, we list the ablation variants in which specific components are removed or modified. Reviewing the three key components described in Alg.~\ref{alg:k.gred}: (i) PGD-based attack initialization, (ii) $k$-based selection of AE candidates, and (iii) $k^*$-driven optimization. We abbreviate these three components as \textit{AE Initia.}, \textit{K-Sel.}, and \textit{Opt.}, respectively, in the following Table~4.

\begin{table}[h]
  \caption{Deatils of PR-based UAP attack and ablation variants.}
  \label{PR UAP candiate}
  \begin{center}
    \begin{small}
      \begin{sc}
        \begin{tabular}{lccc}
          \toprule
          Index \quad \,\, Name      & AE Initia.  & K-Sel.      & Optz.  \\
          \midrule
          (1):\quad\quad\,\,\, FUAP    & FGM(1) & $\times$ & $\times$  \\
          (2): \quad\,\,\,\,\, KFUAP & FGM(N) & $\surd$  & $\times$ \\
          (3):\quad\quad\,\,\,\, PUAP    & PGD(1) & $\times$ & $\times$  \\
          (4): \quad\,\,\,\, KPUAP   & PGD(N) & $\surd$  & $\times$   \\
          (5): \,\,\,PR-based UAP     & PGD(N) & $\surd$  & $\surd$ \\
          \bottomrule
        \end{tabular}
      \end{sc}
    \end{small}
  \end{center}
  \vskip -0.1in
\end{table}
For the attack-initialization component, in addition to the PGD-based initializer in Alg.~\ref{alg:k.gred}, we replace it with a random-start FGM initializer to form an alternative strategy. The resulting variants are denoted as FUAP and KFUAP in Tab.~\ref{PR UAP candiate}. For variants that include AE candidate selection (K-Sel.), the initialization attack is repeated multiple times, denoted by '(N)' after the corresponding attack name. In parallel, variants that select the AE candidate with the largest $k$ are prefixed with 'K'. Otherwise, the sole available AE candidate is used as \textbf{the selected result}. Finally, to validate the effectiveness of the optimization, only the proposed PR-based UAP updates the perturbation using $\nabla_x k^*$, whereas the other variants directly adopt the AE candidate \textbf{selected in the previous step} as the generated perturbation.

\section{Additional Experimental Result}
\label{Additional Experimental Result}
In this section, in addition to FNR and Accuracy results shown in main paper, we provide additional experimental results with a different evaluation metric for the two experiments and the ablation study mentioned in Sec.~\ref{Chapter4:Experimental Result}. Besides, we also report attack performance across multiple intrusion detection datasets and different DRL architectures, transferability evaluation under the black-box setting, and ablation study of epsilon step.
\subsection{F1 score of the results in main paper}

The classifier’s F1 score over the displayed $\epsilon$ range are provided for a more comprehensive assessment of overall performance, which is calculated according to Eq.~\ref{equation:f1}: 
\begin{equation}
\text{F1-score} =
\frac{2 \cdot \text{True Positives}}
{2 \cdot \text{True Positives} + \text{False Positives} + \text{False Negatives}}
\text{.}
\label{equation:f1}
\end{equation}
The F1 score complements accuracy and FNR by offering a more balanced assessment by capturing both false alarms and missed detections in a single metric. This offers an additional evaluation perspective and enables a more comprehensive view of overall detection performance.

\begin{figure*}[h]
\centering
\includegraphics[width=0.99\textwidth,height=0.21\textheight,keepaspectratio]{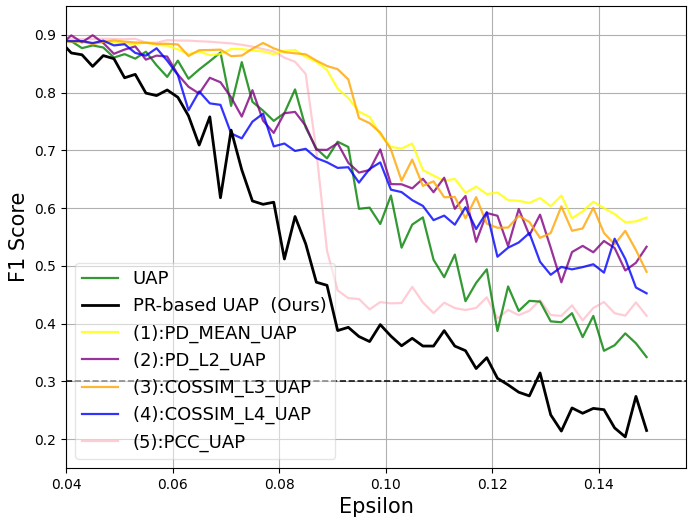}
\caption{F1 score of the proposed PR-based UAP and other UAP baselines}
\label{app Figure Main}
\end{figure*}

Fig.~\ref{app Figure Main} presents the F1 score of the proposed PR-based UAP and other UAP baselines. Consistent with the findings delivered in Sec.~\ref{Chapter4:Experimental Result}, the proposed PR-based UAP achieves the lowest F1 score among all the other baselines reported in Tab.~\ref{tab:multi-loss} over the considered $\epsilon$ range, indicating the strongest degradation of detection performance,i.e., strongest attack effectiveness. Notably, even at $\epsilon=0.09$, where the FNR of the PR-based UAP is the same as that of PCC\_UAP previously introduced in Fig.~\ref{Figure3}(b), the PR-based UAP still attains a lower F1 score, specifically it is the only candidate whose F1 score drops below 0.3 for $\epsilon>0.13$, thereby achiving a stronger attack.

\begin{figure*}[h]
\centering
\includegraphics[width=0.99\textwidth,height=0.21\textheight,keepaspectratio]{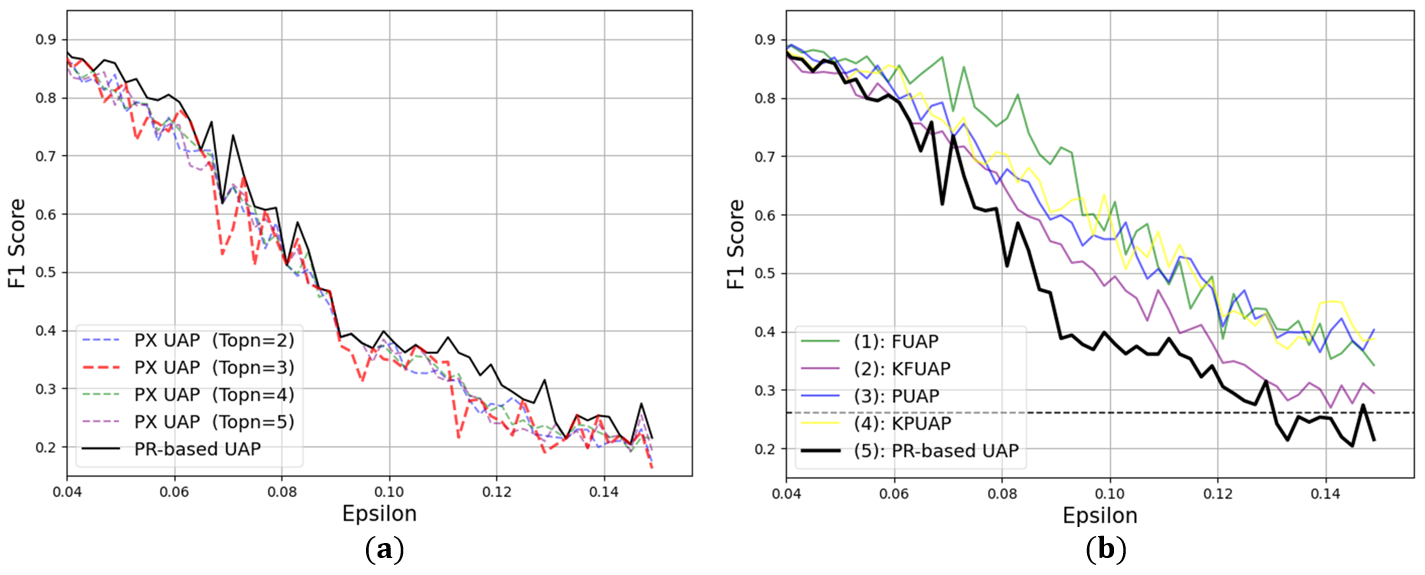}
\caption{F1 score of (a): PX-UAP with different $\mathrm{Topn}$ values. (b): The ablation study}
\label{app Figure XAI}
\end{figure*}

We next report the F1 score of PX-UAP with different $\mathrm{Topn}$ values (Fig.~\ref{app Figure XAI}(a)) and the ablation study (Fig.~\ref{app Figure XAI}(b)). Overall, the trends in both subfigures are consistent with the results shown in main paper. Specifically, PX-UAP yields additional performance degradation in a wide range of perturbation, i.e., $\epsilon<0.07$ and $\epsilon\ge 0.10$. Among all settings, $\mathrm{Topn}=3$ produces the largest reduction in the F1 score. Turning to the ablation study, the PR-based UAP achieves the lowest F1 score relative to all ablation variants, indicating the most pronounced degradation in detection performance and the most effective attack.

In summary, the reported F1-score results provide a complementary evaluation perspective and further substantiate the effectiveness of our proposed UAP methods.

\subsection{Generalizability across Datasets and DRL Architectures}

In the main paper, we report the results of the proposed PR-based UAP on the CICIDS2018 dataset. To further evaluate its generalizability across different scenarios, we additionally launch the proposed PR-based UAP attacks against multiple classifiers with different DRL architectures and on two commonly used network intrusion detection datasets, namely NSL-KDD~\cite{tavallaee2009detailed} and UNSW-NB15~\cite{moustafa2015unsw}.

\begin{figure*}[h]
\centering
\includegraphics[width=0.99\textwidth,height=0.21\textheight,keepaspectratio]{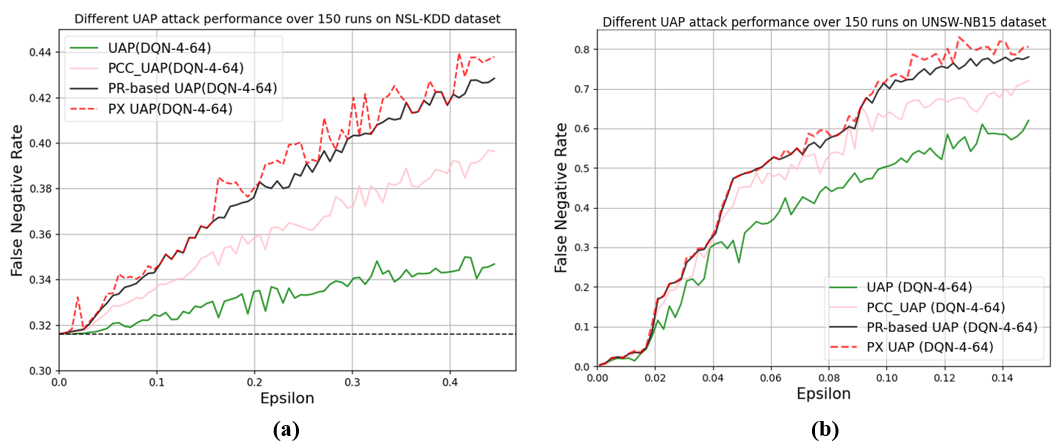}
\caption{FNR of proposed PR-based UAP on DRL-based IDS (DQN-4-64) on (a): NSL-KDD over $\epsilon \in [0, 0.45]$, (b) : UNSW-NB15 over $\epsilon \in [0, 0.15]$.}
\label{app: datasets}
\end{figure*}

Specifically, since the two datasets differ in dimensionality and feature structure, we adopt dataset-specific $\epsilon$ ranges to better demonstrate attack performance, with $\epsilon \in [0, 0.45]$ for NSL-KDD and $\epsilon \in [0, 0.15]$ for UNSW-NB15, resceptively. The data preprocessing follows the same procedure described in Section~\ref{Experimental Settings}. It can be shown in Fig.~\ref{app: datasets} that the proposed PR-based UAP attack and PX-UAP attack consistently outperform the state-of-the-art PCC\_UAP on the other two datasets with FNR $\approx$ 0.43 and 0.78, respectively, demonstrating the generalizability of our proposed method in different scenarios in the intrusion detection area.

Furthermore, in addition to the DRL-based IDS architecture as the target classifier as used in the main paper, we further assess the attack performance on the same CICIDS2018 dataset but under different DRL agents and network architectures to assess whether the proposed attack can consistently compromise diverse DRL-based IDS models. It should be noted that the notation in parentheses indicates the DRL agent and network architecture used by the target IDS. For example, DQN-6-128 denotes a DQN-based target model with six hidden layers, each containing 128 units.

\begin{figure*}[h]
\centering
\includegraphics[width=0.99\textwidth,height=0.21\textheight,keepaspectratio]{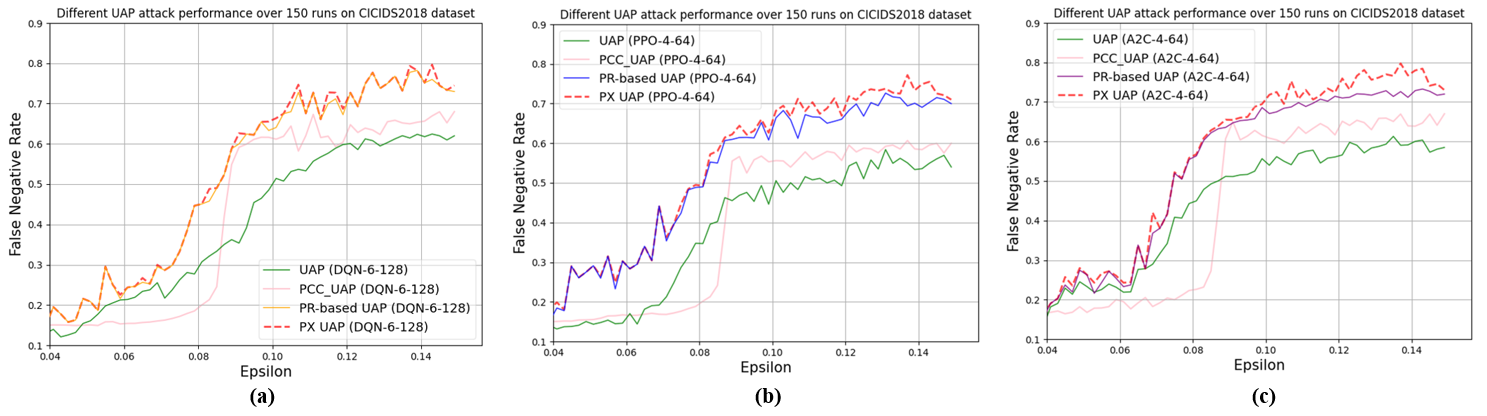}
\caption{FNR of proposed PR-based UAP on different DRL-based IDS on CICIDS2018 dateset with the taget model of (a): DQN-6-128, (b): PPO-4-64, (c): A2C-4-64.}
\label{app: different DRL}
\end{figure*}

As shown in Fig.~\ref{app: different DRL}, in addition to DQN, we further consider two widely used DRL architectures (agents): Proximal Policy Optimization (PPO)~\cite{schulman2017proximal} and Advantage Actor-Critic (A2C)~\cite{mnih2016asynchronous}. Specifically, the three target models used for evaluation are DQN-6-128, PPO-4-64, and A2C-4-64. It can be observed that the proposed PR-based UAP consistently outperforms PCC\_UAP across all three target models and PX-UAP further improves the attack performance based on that, further demonstrating the generalizability of our method under different intrusion detection scenarios and the effectiveness of our algorithmic design.
\subsection{Transferability Evaluation under the Black-Box Setting}

\begin{figure*}[h]
\centering
\includegraphics[width=0.99\textwidth,height=0.21\textheight,keepaspectratio]{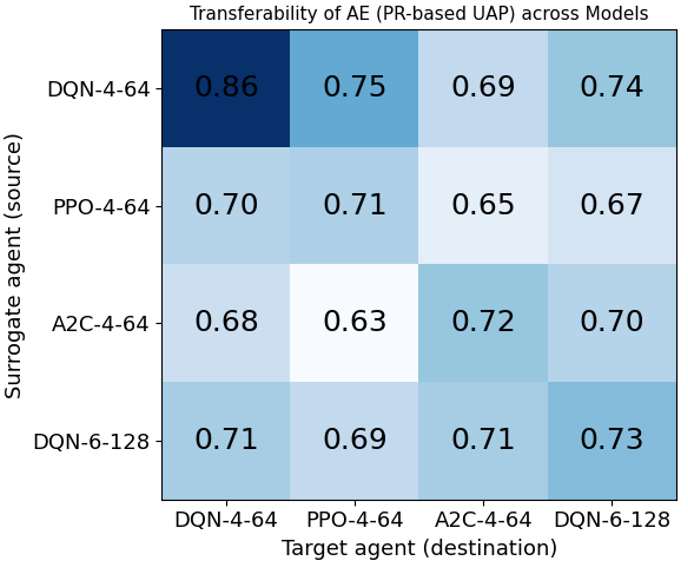}
\caption{FNR for Cross-model Transferability of AE (PR-based UAP) among different DRL-based IDS model on CICIDS2018 at $\epsilon = 0.15$}
\label{Re_Figure:Transferablities}
\end{figure*}

In some real-world cybersecurity scenarios, attackers may not have access to the internal structure or parameters of the target model, corresponding to a black-box attack setting. Therefore, this setting should also be considered for a comprehensive evaluation of attack effectiveness. We conduct additional transferability experiments, where UAPs are crafted on a surrogate model with white-box access and then applied to other target models to evaluate their transferability. The four models used as the surrogate and target models are those considered in the previous CICIDS2018 evaluation in the above section, together with the model used in the main paper, i.e., DQN-4-64.

The results shown in Fig.~\ref{Re_Figure:Transferablities} show the performance of the generated AEs under different surrogate-target model pairs. It can be observed that, across different transfer scenarios, i.e., the off-diagonal entries, the proposed PR-based UAP still achieves decent attack performance, which provides evidence of the proposed attack's effectiveness even beyond the original white-box setting, as well as the transferability of the generated AEs across different models and architectures.

\subsection{Ablation study of epsilon step}
In this section, we conduct an ablation study on the setting of $\epsilon$-steps to verify that the strong effectiveness of our proposed attack stems from the attack design itself, rather than being attributable to favorable $\epsilon$-step selection. The results demonstrate that our proposed PR-based UAP consistently outperforms the two baselines, both under the fixed unified $\epsilon$-step setting and when $\epsilon$-steps are tuned for each method.

Specifically, in the experimental setting of Alg.~\ref{alg:uap} in the main paper, we set the number of $\epsilon$-steps to 1, which means each update performs a full-$\epsilon$ step at line~7. This design choice is intended to ensure a unified fairness criterion, as different UAP methods may benefit from different small or adaptive step-size schedules. If each candidate were allowed to use its own step-size schedule, the performance comparison would be affected by step-size tuning as well, rather than reflecting only the difference in UAP design. Therefore, we adopt the same full-$\epsilon$ update rule across all methods in the experiments shown in Chapter~\ref{Chapter4:Experimental Result} to ensure a fair comparison.

Despite this setting, in order to provide a more comprehensive evaluation of our proposed attack method and rule out the influence of $\epsilon$-step selection, we further conduct an additional step-size sensitivity analysis for all attack candidates. Namely, we compare the attack performance of the proposed PR-based UAP, PCC\_UAP, and UAP under different step sizes, identify the best-performing step size for each candidate method, and finally report a best-step-size comparison.

\begin{figure*}[h]
\centering
\includegraphics[width=0.99\textwidth,height=0.21\textheight,keepaspectratio]{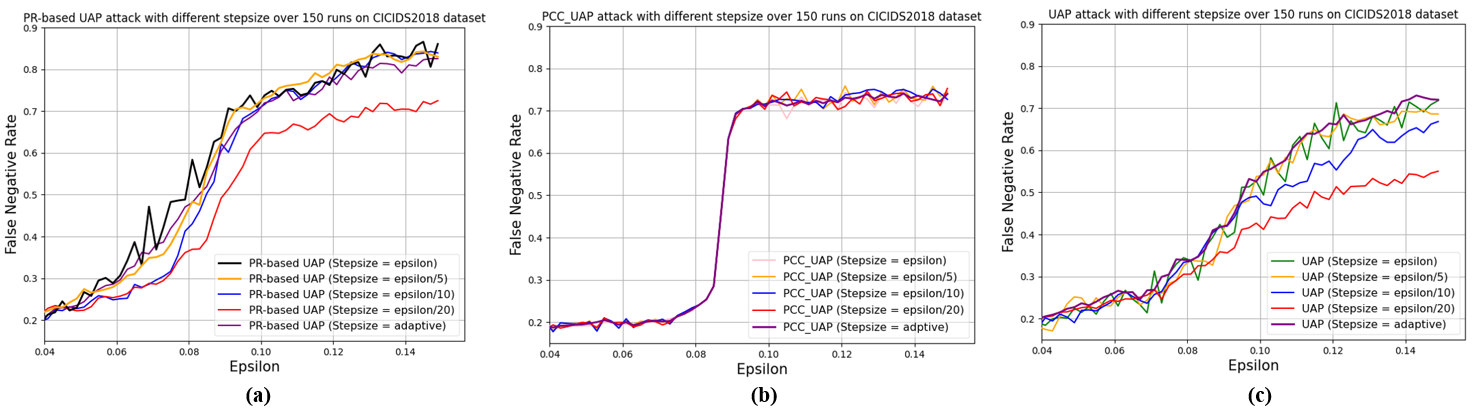}
\caption{FNR of (a): PR-based UAP (b): PCC\_UAP (c): UAP with different stepsize on CICIDS2018 dateset. The applied stepsize are show in parentheses.}
\label{Re_Figure:attack epsilon step}
\end{figure*}

Fig.~\ref{Re_Figure:attack epsilon step} show attack performance of PR-based UAP, PCC\_UAP, UAP under different $\epsilon$ stepsize. We set the step size to $\epsilon$ (the original setting), $\epsilon$/5, $\epsilon$/10, $\epsilon$/20, and an adaptive step-size scheme. In the adaptive setting, the step size is gradually reduced over the 30 iterations (Alg.~\ref{alg:uap} line~2): $\epsilon$/5 for iterations 1–10, $\epsilon$/10 for iterations 11–20, and $\epsilon$/20 for iterations 21–30. As discussed above, different attack methods may benefit from different step-size schedules. Therefore, according to the results in Fig.~\ref{Re_Figure:attack epsilon step}(a)--(c), we select the best-performing step-size setting for each attack method, namely UAP (adaptive) and PCC\_UAP ($\epsilon/10$) and then compare them with two comparably good settings of the proposed PR-based UAP (Fig.~\ref{Re_Figure:attack epsilon step}(a)), including $\epsilon/5$ and the original full-$\epsilon$ step setting.

\begin{figure*}[h]
\centering
\includegraphics[width=0.99\textwidth,height=0.21\textheight,keepaspectratio]{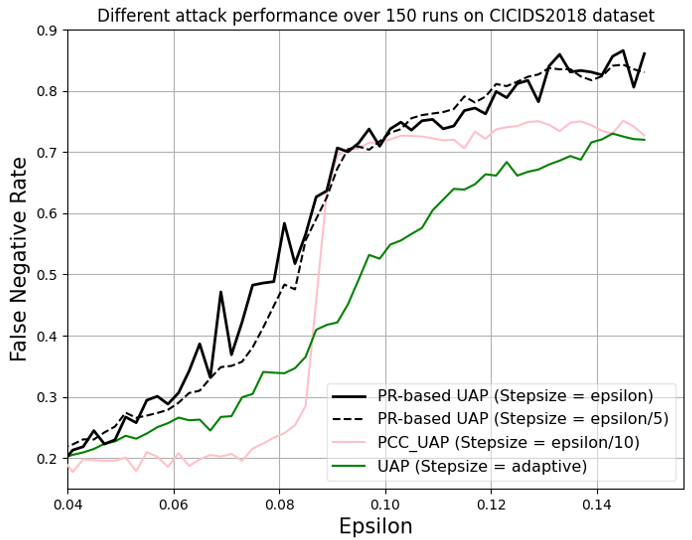}
\caption{FNR of best-step-size comparison for PR-based UAP, PCC\_UAP and UAP attack.}
\label{Re_Figure:Stepsize compare}
\end{figure*}

As shown in Fig.~\ref{Re_Figure:Stepsize compare}, the two PR-based UAP variants (black solid and dashed lines) still outperform the other two baselines under their respective best-performing settings. Together with the original comparison under a unified full-$\epsilon$ step size, this supports a more comprehensive conclusion that the proposed PR-based UAP significantly outperforms the state-of-the-art PCC\_UAP under both a fairness setting with the same step size and a fairness setting with the best step size, which further demonstrates the effectiveness of our proposed attack method.

\section{Limitation and future work}
In our opinion, the limitation of this work and the corresponding directions for the future work are as follows:
Although this work evaluates the transferability of the proposed attacks to provide evidence of their effectiveness under black-box settings, our main threat model remains white-box. Future work can further extend this study to more restrictive black-box scenarios. This would enable a closer approximation of practical deployment environments.
\section{Experiment Device}
All experiments in this study were conducted on a laptop workstation equipped with an AMD Ryzen 9 5900HX processor, an NVIDIA GeForce RTX 3080 Laptop GPU with 16 GB VRAM, 32 GB DDR4 RAM, and 2 TB PCIe Gen3 SSD storage. The operating system used throughout the experiments was Windows 10 64-bit.
\section{Broader Impact}
This work may have both positive and negative societal impacts. Positively, by exposing the vulnerability of DRL-based intrusion detection systems to universal adversarial perturbations, this study can help the research community and security practitioners better understand potential weaknesses in current IDS models and motivate the development of more robust and reliable cybersecurity defenses. This is particularly important for improving the trustworthiness of intelligent security systems in practical deployment. Negatively, since the work studies adversarial attack methods, the proposed techniques could potentially be misused to evade real-world intrusion detection systems. Therefore, the study is conducted in controlled offline experimental settings and is intended to support academic research and defensive security development.


\end{document}